\documentclass{article}

    \usepackage[preprint, nonatbib]{neurips_2019}

\usepackage[utf8]{inputenc} 
\usepackage[T1]{fontenc}    
\usepackage{hyperref}       
\usepackage{url}            
\usepackage{booktabs}       
\usepackage{amsfonts}       
\usepackage{nicefrac}       
\usepackage{microtype}      
\usepackage[pdftex]{graphicx} 
\usepackage{amssymb,amsmath}
\DeclareMathOperator{\E}{\mathbb{E}}

\def\bm{{\bf m}}

\def\bz{{\bf z}}

\def\bv{{\bf v}}
\def\bu{{\bf u}}
\def\bs{{\bf s}}

\def\bd{{\bf d}}

\def\mR{\mathcal{R}}

\def\pprior{\mathcal{P_{\rm pr}}}

\title{Generative Learning of Counterfactual for Synthetic Control Applications in Econometrics}

\author{%
  Chirag Modi \\ 
  Department of Physics\\
  University of California, Berkeley\\
  \texttt{modichirag@berkeley.edu} \\
   \And
   Uro\v{s} Seljak \\
  Department of Physics\\
  University of California, Berkeley\\
   \texttt{useljak@berkeley.edu} \\
}

\begin{document}

\maketitle

\begin{abstract}
A common statistical problem in econometrics is to estimate the impact of a treatment
on a treated unit given a control sample with untreated outcomes.
Here we develop a generative learning 
approach to this problem, learning the probability distribution of the data, which can 
be used for downstream tasks such as post-treatment counterfactual prediction and hypothesis testing. 
We use control samples to transform the data to a Gaussian and 
homoschedastic
form and then perform Gaussian process analysis in Fourier space, evaluating the 
optimal Gaussian kernel via non-parametric 
power spectrum estimation. We combine this Gaussian 
prior with the data likelihood given by the pre-treatment data of the single unit, to obtain the synthetic prediction of the 
unit post-treatment, which minimizes the error variance of synthetic prediction. 
Given the generative model the minimum variance counterfactual is unique, and comes with an
associated error covariance matrix. 
We extend this basic formalism to 
include correlations of primary variable with other covariates of interest. Given 
the probabilistic description of generative model we can 
compare synthetic data prediction with real data
to address the 
question of whether the treatment had a statistically significant impact. For this purpose  
we develop a hypothesis testing approach and evaluate
the  Bayes factor.  
We apply the method to the well studied example of California (CA) tobacco sales tax of 1988. 
We also perform a placebo analysis using control states to validate our methodology. Our hypothesis testing method suggests 5.8:1 odds in favor of CA tobacco sales tax having an impact on the tobacco sales, a value that 
is at least three times higher than any of the 38 control states.  
\end{abstract}

\section{Introduction}
\label{sec:Introduction}
Drawing inferences about the causal effect of policy interventions from observational data is a challenging problem in econometrics.
Given that one cannot observe the causal effects directly, the estimates of them are ultimately
based on comparisons of different units with different levels of such interventions.
In last few decades, several works have developed regression analysis, difference in difference,
and more recently popularized synthetic control methods (SCM), to tackle this problem \cite{Athey2016, Abadie03, Abadie10}.
These approaches measure the difference between the predicted synthetic post-treatment data
for the treated unit ({\it counterfactual}) and the observed outcomes to estimate the impact.

A limitation of all these methods is the inability to achieve exact balance (i.e. fit/reproduce exactly) on the
pre-treatment outcomes for the treated unit, which can lead to significant biases \cite{Ferman16, Ben-Michael18, synetheticdid}.
Additionally, there is also no first principled way of doing a secondary analysis which would shed light on the credibility of the primary analysis,
quantifying the statistical significance of the observed difference, and as a result several forms exist in the literature \cite{Athey2016, Doudchenko16}.

In this work, we propose a different approach to do causal inference, by performing a longitudinal time correlation analysis to learn the generative model of the data from the control samples. With a 
generative model we manage to achieve exact balance on the pre-treatment data and predict the likelihood of the post-treatment data, thus allowing one to develop
principled ways of doing a secondary analysis.

\section{Methodology}
\label{sec:Methodology}
Our approach is to learn the generative model by performing a longitudinal time correlation analysis, which is especially suited to panel data studies in econometrics.
We begin with the time series data ($\bd_{i}$; i$^{th}$ unit) after removing the global mean. 
Next we transform the data into approximately Gaussian  and homoschedastic form with a 
nonlinear transformation $\bz_i = \Psi(\bd_i)$ \cite{boxcox, Seljak19}).
Our generative model for these transformed data is multi-variate Gaussian, with covariance matrix that depends only on relative time difference.
We do a non-parametric Gaussian process (GP) analysis using Fourier transformation, i.e. 
instead of doing a GP analysis on the time-series data directly, we work with their Fourier modes ($\bs_i = \bu_i + i\bv_i$),
estimated by Fourier transforming ($\mathcal{R}^\dagger$) the Gaussian data i.e. $\bs_{i} = \mathcal{R}^\dagger\bz_i=\mathcal{R}^\dagger\Psi(\bd_i)$, i.e. 
$ s_{i\nu} = \sum_{t=1}^T e^{-i \frac{2\pi}{T}\nu t} \Psi(d_{it})$, where $\nu$ is the Fourier mode 
frequency. 

These Fourier modes allow us to learn the underlying correlations in data on different time-scales.
One can use the control unis to estimate the Gaussian prior on the Fourier modes by measuring the power spectrum ($\mathcal{P}_{\rm pr}(\nu)$).  
This thus amounts to learning the GP kernel for the temporal fluctuations of the data in a non-parametric fashion.


To estimate the Fourier modes for the treated unit of interest (say unit `0'), the Gaussian prior (kernel) power spectrum is
combined with the data likelihood, which is Gaussian. This gives the posterior 
\begin{equation}
    P(\bs_0 | \bd_0) \propto {\rm exp} \Big(-\frac{1}{2} \Big\{\bs_0^\dagger \pprior^{-1}\bs_0 + \
        [\Psi(\bd_0) - \mathcal{R}\bs_0]^\dagger \mathcal{N}^{-1}[\Psi(\bd_0) - \mathcal{R} \bs_0] \Big\}\Big)
    \label{eq:posterior1}
\end{equation}

where $\mathcal{N}$ is the noise-covariance matrix for the observed data. 
For the likelihood of the Gaussianized data, the noise variance for the pre-treatment period ($\sigma_t$) can be vanishingly small, enabling us to achieving exact balance on these points.
One can maximize this posterior with respect to the Fourier modes
in order to get the maximum-a-posteriori (MAP) estimate of these modes.
Ignoring all the irrelevant constants, this amounts to minimizing the negative log-posterior:

\begin{equation}
    \mathcal{L} = \sum_{t = T_i=1}^{T_0} \frac{ \Big((\mathcal{R} s_0)_t - \Psi(d_0)_t \Big)^2}{\sigma_t^2} + \sum_{\nu}\frac{\bs_{0\nu}\bs_{0\nu}^*}{\pprior(\nu)},
    \label{eq:posterior2}
\end{equation}
where $\sigma_t \rightarrow 0$.
This procedure is known as Wiener filter analysis \cite{wiener64}
and it minimizes the error variance of synthetic prediction \cite{rybicki92}, thus resulting in the optimal synthetic prediction of the unit post-treatment.

Transforming the data to a Gaussian form 
is desirable since the counterfactual can be shown to be minimal variance for Gaussian data, and the prior on the Fourier modes is entirely described by its
covariance matrix. 
Homoscedastic form ensures the Fourier modes are 
uncorrelated, 
reducing the covariance matrix in Fourier space from full-rank to diagonal,
thus massively reducing the number of parameters 
that need to be extracted from the data. 
We achieve Gaussianization by using a series of bijective non-linear (arcsinh and Yeo-Johnson) transformations ($\Psi$),
the parameters for which are fit by maximizing the likelihood of the observed control units data. 

\subsection{Covariates}
An additional challenge in econometrics is to correctly handle covariates, i.e. other variables that can influence
the primary outcomes irrespective of intervention.
In our framework, its straightforward to include these covariates through their cross-correlation (cross-spectra) with the primary variable. 
Thus we simply concatenate the the data and the Fourier mode vectors for these two variable ($\bs_0^{ab} = (\bs_0^{a \frown} \bs_0^b)$)
and learn the prior covariance matrix on these modes from the control units.
The covariance matrix now consists of the mean auto-spectra for each data variable on the diagonal as well as the mean cross-spectra (cross-correlation)
$ \mathcal{P}_{i}^{ab} = \bs^a_i \bs_i^{b\dagger}$ on the major block off-diagonal.

\section{Secondary analysis : hypothesis testing}
\label{sec:secondary}
A novel feature of our analysis is that in addition to predicting a mean counterfactual observation (point estimate),
we also get a covariance matrix for the synthetic prediction 
from the inverse Hessian of the Gaussian posterior (Eq. \ref{eq:posterior1}) at the MAP,
\begin{equation}
    C_{\bs} = \E[(\bs-\hat{\bs})^\dagger (\bs-\hat{\bs})] = \left( \nabla_{\bs}\nabla_{\bs} \mathcal{L}\right)^{-1}= (\pprior^{-1} + \mR^\dagger \mathcal{N}^{-1} \mR)^{-1}
    \label{eq:cov}
\end{equation}
This is the covariance in Fourier space and correspondingly for the predicted model in data space (\bz)
\begin{equation}
    C _\bz= \mR((\pprior^{-1} + \mR^\dagger \mathcal{N}^{-1} \mR)^{-1}\mR^\dagger 
    \label{eq:cov2}
\end{equation}

This matrix is in general non-diagonal. 
This covariance matrix allows us to evaluate the likelihood of the observed post-treatment data for different synthetic predictions 
and develop principled ways to do a consistent secondary analysis to measure the statistical significance of any observed impact.
We investigate this for two different approaches, performing hypothesis testing with a a-posteriori and a-priori model
and elucidate the pitfalls of the former. 

\subsection{a-posteriori method : upper limit}
The simplest hypothesis test is to do a simple likelihood ratio test,
where we compare the likelihood of the observed data assuming different counter-factuals for a fixed covariance.
To model this, we parameterize our model prediction with parameter $\alpha$, which allows us to smoothly interpolate
the counter-factual between our counterfactual model prediction ($\hat \bd^m_{0}$) and the actual observed data ($\bd^I_{0}$).
\begin{equation}
    m_t(\alpha) = \alpha (d^I_{0, t} - \hat d^m_{0, t}) +  \hat d^m_{0, t} \quad \forall  t \in (T_0, T_f).
    \label{eq:model1d}
\end{equation}

We define the null 
hypothesis A as the treatment having no effect, which corresponds to $\alpha=0$ since $[\bd^I_{0}|\bm(0)]$ measures the likelihood of
the observed post intervention data under the modeled counter-factual. 
We define the alternative hypothesis B as corresponding to $\alpha=1$,
where the likelihood ratio is maximized ($p_B=1$).
Then, the likelihood ratio test, after transforming to Gaussian space ($\bz$) and taking corresponding Jacobian into account is
\begin{equation}
\frac{p_B}{p_A}<
    \exp\left[\big(\hat \bz^m_{0} - \bz^I_{0}\big)^T C^{-1} \big(\hat \bz^m_{0} - \bz^I_{0}\big)/2\right] = \exp\left(\chi^2/2\right)
    \label{eq:upper}
\end{equation}

This is an upper-limit to the alternative hypothesis and can be 
quite unreasonable, since it assumes that the model can exactly fit the observed data
at $\alpha=1$, which is maximally a posteriori approach.
Moreover, there is no penalty for having an extra parameter $\alpha$.

\subsection{a-priori method with Bayes penalty}

To perform a-priori hypothesis testing where our models are not influenced by the post-treatment data of the treated unit sample, we modify the 
aforementioned hypothesis B. We wish to have some parametrized model that 
allows post-treatment prediction to differ from $\hat{d}_{0, t}^m$, and 
since we are testing a specific hypothesis we impose additional constraints on 
the model, such as $m_t < \hat{d}_{0, t}^m$. This correction has to vanish 
at the time of treatment $T_0$. A simple form 
is $n^{\rm th}$ order polynomial correction in the prediction to improve our model for the counter-factual.
Thus, for $n=2$, our model becomes 
\begin{equation}
    m_t(\alpha,\beta) = \hat{d}_{0, t}^m  + \alpha (T-T_0) + \beta (T-T_0)^2 \quad \forall  t \in (T_0, T_f),
    \label{eq:model2d}
\end{equation}

The two new parameters, $\alpha$ and $\beta$, increase the flexibility of the model but do not fit the observations exactly.
To perform hypothesis testing we marginalize over these two parameters over their prior.
This is akin to Bayes/Occam's razor penalty.
To estimate the prior on these parameters ($p(\alpha, \beta)$) needed for marginalization, while keeping our model non-parametric and conservative,
we estimate a flat prior from the control units, together with the 
condition of the hypothesis B (e.g. that the value is reduced relative to the 
counterfactual).
Then, our hypothesis A again corresponds to the observed data being likely under the predicted counter-factual 
\begin{equation}
p_A(\bd^I_{0})=p[\bd^I_{0}|\bm(\alpha=0,\beta=0)).
\end{equation}
however hypothesis B is the data likelihood averaged over $\alpha$ and $\beta$: 
\begin{equation}
    p_B(\bd^I_{0})=\int d\alpha d\beta p(\alpha,\beta) p[\bd^I_{0}|\bm(\alpha, \beta)]. 
    \label{eq:marginalpB}
\end{equation}

\section{An example: California tobacco tax of 1988}
\label{sec:supplementary}
As an example we apply our formalism to one of the well studied problems of econometrics.
The state of California (CA) implemented tax on Tobacco sales in 1998. 
Given the outcomes like tobacco sales for CA and all other states over the period of time pre and post-intervention (1970-2017), 
we are interested in investigating if the tax had any impact on the tobacco sales in the state. 

\begin{figure}
  \centering
   \includegraphics[width=1.0\linewidth]{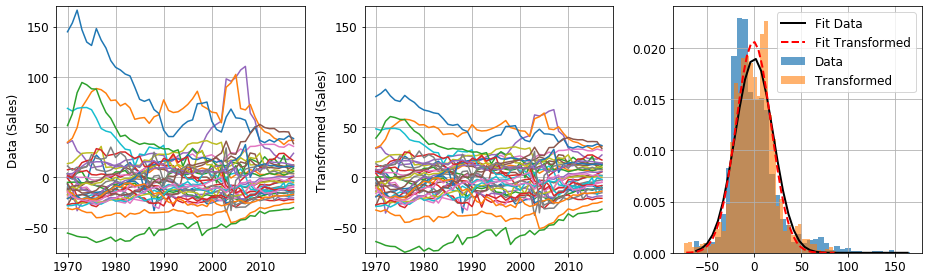}
  \caption{Gaussianizing the data : (Left) Time series data for tobacco sales per capita after removing national mean for all the states (Center) Time series data after Gaussianizing with bijective non-linear transforms. Outliers are suppressed and data becomes more homoschedastic. (Right) Corresponding data histograms and Gaussian fits to them before and after the non-linear transforms}
  \label{fig:transform}
\end{figure}

\begin{figure}
  \centering
   \includegraphics[width=1.0\linewidth]{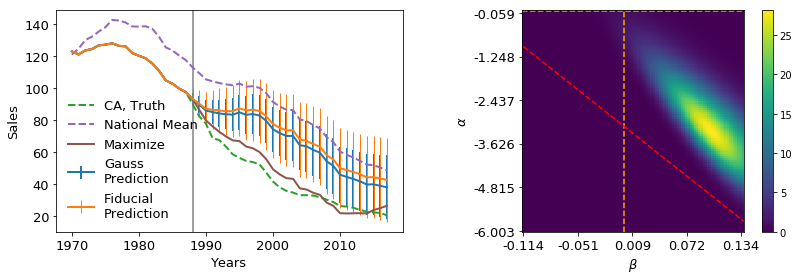}
  \caption{(Left) Counterfactual prediction for tobacco sales for CA with and without Gaussianizing data (bue vs orange) with corresponding errors bars. For comparison, 
  also shown are true sales (dashed green) and national mean (dashed purple).
  The solid brown line is the maximum likelihood line for a-priori hypothesis testing method.
  (Right) We show the likelihood of the data for a-priori model as a function of the two parameters, 
  normalized with the value at the predicted counter-factual ($\alpha, \beta=0$, dashed orange lines).
  The region below the dashed red line is excluded based on the physical arguments, 
  by constraining the model to not predict non-negative sales.
  }
  \label{fig:maximize2d}
\end{figure}

We perform our analysis on the tobacco sales data using other 38 states of US (after eliminating states with their own version of tobacco tax) \cite{sourcetobacco, Orzechowski12}. Fig. \ref{fig:transform}
shows the time series data for these states before and after Gaussianizing with the bijective non-linear transformations. This suppresses the outliers as well makes the data more homoschedastic. Fig. \ref{fig:maximize2d} shows the result of our analysis.
As expected, our model is able to fit the California sales in pre-intervention years exactly, unlike other regression and synthetic model studies \cite{Ben-Michael18}.
Immediately after the intervention ($T_0$), we see that the counter-factual California data is still driven by the long-range temporal correlations learned from fitting the pre-intervention years.
As we move away from intervention, the model is increasingly driven by the mean of 
control states and hence approaches the national mean, as one would expect in the absence of any other extraneous information.
Moreover, the error in the model prediction increases as we move further away from $T_0$, which is expected as the influence of the pre-treatment data decreases. 
The errors are also larger for fiducial analysis than the Gaussian analysis since the latter suppresses the outliers. Given our model prediction, we estimate a decrease of 34 packs in sales in California in 2000 due to tax imposed in 1988. 

To estimate the significance of this decrease, we do both the a-priori and a-posteriori hypothesis tests.
For the a-posteriori method, we find an upper bound on likelihood ratio of $\sim 2000$.
This is unreasonably high which is due to the alternative hypothesis fitting the observed data exactly.
Performing a placebo analysis, where we exchange CA with other control states and repeat the analysis,
finds similarly high or even higher evidence in favor of intervention for other states without intervention, thus highlighting that this methodology is not robust. 

The results of the a-posteriori hypothesis testing with Bayes penalty are shown in Fig. \ref{fig:maximize2d}.
The right panel shows the likelihood ratio with the fiducial counterfactual prediction ($\alpha, \beta=0$),
as a function of the polynomial parameters ($\alpha, \beta$). In the left figure, in brown line, 
we show the model which maximizes this ratio and note that it does not fit the observed data exactly, thus avoiding over-fitting.
The flat prior on $\alpha, \beta$ is chosen by fitting 38 control states with the polynomial, thus being maximally conservative.
This prior space is constrained by physical arguments, such as rejecting the space that leads to negative sales (below red line).
We find the Bayes factor after marginalization to be in favor of hypothesis B (tax having an impact) with odds $5.8:1$. 
This is more conservative than simply quoting the ratio with the maximum $\alpha, \beta$, which gives likelihood ratio of $\sim28$.
While this number should still be viewed as a guidance, since it does depend on the choice of the prior for $\alpha$ and $\beta$, 
placebo analysis confirms that its more robust than the a-posteriori method. The 5.8 evidence in favor of CA tax having an impact is at least 3 times higher than for any of the 38 other control states, most of which have 
Bayes factor less than one, as expected for states without intervention. These results
suggest that there is evidence that the 
sales tax in CA had an impact on the tobacco sales.

\begin{figure}
  \centering
   \includegraphics[width=1.0\linewidth]{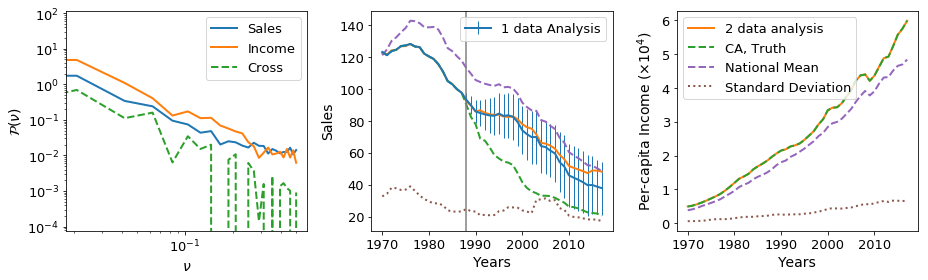}
\caption{Covariate Analysis: (Left) We show the mean auto and cross spectra for per capita tobacco sales and personal income. (Middle) We show the predicted sales for the counterfactual CA with personal income covariate (2 data, orange) and compare it with the predicted sales of the 1 data analysis without the 
    covariate (blue). In addition we show the national mean sales, their variance and true CA sales. (Right) We show true CA personal income as well as the fit (orange) from our analysis. Again, we also show the national mean and variance for comparison.}
  \label{fig:covariate}
\end{figure}

To demonstrate our methodology for multiple datasets, we use personal income as another covariate \cite{sourceincome}. Fig. \ref{fig:covariate} shows the result for this analysis. We show the auto and cross spectra for the sales and income. Since the cross-spectrum is lower than the geometric mean of the auto-spectra, these variables are not very highly 
correlated, although there is some correlation. This is reflected in the middle panel where we find that the counterfactual prediction in this case does not differ much from our single dataset analysis. In the right panel, we show that for personal income, as with the pre-intervention tobacco sales data, we are able to achieve perfect balance.

\section{Conclusion}

We present generative learning approach to causal inference based on modeling the probability distribution of data as a Gaussian distribution, after transforming the data into a  Gaussian and homoscedastic form. The benefits of such analysis are that it is unique, and that 
it gives probabilistic description of the data, allowing subsequent secondary 
analysis such as counterfactual prediction and hypothesis testing. Placebo analysis confirms the statistical power and robustness of the method. We hope that the qualities of this approach will be validated by applying to other problems of causal inference in econometrics and beyond. 

\newpage
\bibliographystyle{unsrt}  
\bibliography{references}  

\end{document}